# A comparative study between proposed Hyper Kurtosis based Modified Duo-Histogram Equalization (HKMDHE) and Contrast Limited Adaptive Histogram Equalization (CLAHE) for Contrast Enhancement Purpose of Low Contrast Human Brain CT scan images

Sabyasachi Mukhopadhyay[1], Soham Mandal[2], Sawon Pratiher[3], Satyasaran Changdar[2], Ritwik Burman[1], Nirmalya Ghosh[1], Prasanta K Panigrahi[1]
[1]Indian Institute of Science Education and Research, Kolkata
[2]Institute of Engineering & Management, Kolkata
[3]Indian Institute of Technology, Kanpur

*Abstract-* In this paper, a comparative study between proposed hyper kurtosis based modified duo-histogram equalization (HKMDHE) algorithm and contrast limited adaptive histogram enhancement (CLAHE) has been presented for the implementation of contrast enhancement and brightness preservation of low contrast human brain CT scan images. In HKMDHE algorithm, contrast enhancement is done on the hyper-kurtosis based application. The results are very promising of proposed HKMDHE technique with improved PSNR values and lesser AMMBE values than CLAHE technique.

## I. INTRODUCTION

There are two most widely used medical imaging procedures, such as, MRI (Magnetic Resonance Imaging) and CT (Computerized Tomography) scan [9]. It can be mentioned that the contrast enhancement power of global HE method is relatively less optimized. Though local HE can enhance overall contrast more effectively than global HE in specific cases, it effects computation cost is much higher [2-6]. A new grey-level mapping approach for direct histogram equalization was proposed by Y. J. Zhan et.al. Using this method, the accuracy of mapping the original histogram to the specified histogram gets improved [4]. A novel extension of histogram equalization to utilize independent histogram equalizations separately over two sub-images obtained by decomposing the input image based on its mean with a constraint that the resulting equalized sub-images are bounded by each other around the input mean by Y.T. Kim et.al. It was shown mathematically that their proposed algorithm preserves the mean brightness of a given image significantly as well as compared to typical histogram equalization while enhancing the contrast [8]. Here it can be mentioned that the biomedical images are mainly anti-symmetric [16-17]. In this paper, we are dealing with CT scan images of human brain. Mukhopadhyay et.al., already have showed that the CT scan images of human brain are anti-symmetric [14]. Here a comparative study of HKMDHE and CLAHE methods is performed for contrast enhancement purpose for large number of image samples. The improved PSNR values clearly show our proposed HKMDHE technique provides better results. The lesser value of AMMBE clearly reveals that our proposed technique is preserving the brightness. The detail of the proposed technique and results are discussed thoroughly below.

## II. THEORETICAL BACKGROUND

Image histograms are widely used for the contrast enhancement purpose.

## II. (a) Histogram Equalization

If intensity levels are normalized within range of [0, 1] and the probability density function of given input image is $p_r(w)$, then the output intensity level will be, $s = \int_0^r p_r(w)dw$, where, $w$, is a dummy variable of integration.

The probability density function of the output image will be $p_s(s) = 1$ for $0 \leq s \leq 1$ else $p_s(s) = 0$ outside the range of [0, 1]. Therefore the intensities of the output images are equalized within the range of [0, 1], which expands the dynamic range. Hence contrast of the images is enhanced [1, 11].

## II. (b) Contrast Limited Adaptive Histogram Equalization (CLAHE)

There are several techniques for Histogram equalization. Among them contrast limited adaptive histogram equalization (CLAHE) is a very efficient approach. Unlike the conventional technique of operating over whole images, this method processes the small region of images i.e., tiles and combines them with bilinear interpolation to avoid an artifact [10, 12-13].

## II. (c) Formulation of Modified Mean for Hyper-Kurtosis ($\beta$) $\geq \tau$

Kurtosis describes the peak width and tail weight of a distribution. But general kurtosis measures heavy tails; not the peakedness. Therefore to avoid that hyper kurtosis ($\beta$) is computed for providing the comparison with normal distribution. The Power-Law Transformation of mean (m) is applied here due to low contrast nature of the images near the mean. The mathematical expression of Power-Law Transformation is: $M = m^\gamma$, Where, $0 \leq \gamma \leq 1$ and $\tau$ is a threshold value. Different modified means can be obtained by changing the value of $\gamma$. On the basis of this modified mean, the histogram of input image can be separated. A favourable value can be chosen during analysis.

## II. (d) Formulation of Modified Mean for Hyper-Kurtosis ($\beta$) $< \tau$

The modified mean can be obtained as:

$$MM = \sqrt{m \pm \beta}, \quad \beta = \frac{E(X-m)^6}{\sigma} \quad \text{...............} (1)$$

Here 'mean' is the average pixel value of an image. The Positive sign is given when kurtosis is negative otherwise it is negative. $\sigma$ is the standard deviation of the distribution of $X$ which denotes the intensity.

## II. (e) Application of Modified Algorithm

The value of the modified mean obtained from the histogram of the input image is used in the power-law formula depending on the value of hyper-kurtosis ($\beta$). It then divides the histogram of input image into two parts based on this modified mean. Thereafter, HE operation is applied on each segment separately.

$A_t = M_1 \times (CDF_1(B_t))$, here $B_t$ is the $t^{th}$ gray level value.

$$CDF_1(A_t) = \frac{\sum_{n=0}^{t} N_n}{\sum_{p=0}^{M_1} N_p}, \quad N_i \text{ is the number of pixels in the image with gray level } B_i \quad (2)$$

Here $t = 0, 1 \ldots M_1$ and $M_1$ is the modified mean.

$$M_1 = \sum_{t=0}^{M-1} PMF(B_t) \times (B_t) \quad (3)$$

Gray levels above the Modified Mean are pointed to the new gray levels $A_t$. This is shown in eq (4).

$$(CDF_2(B_t)) = \frac{\sum_{n=M_1+1}^{t} N_n}{\sum_{p=M_1+1}^{M-1} N_p} \qquad (4)$$

$A_t = (M-1-M_1) \times (CDF_2(B_t)) + M_1$, Where $t = (M_1+1)...(M-1)$

**II. (f) Peak Signal to Noise Ratio (PSNR)**

For an input image $X(i,j)$ of size $M \times N$, the output image $Y(i,j)$ of same size will produce PSNR value as:

$$PSNR(dB) = 20\log_{10}\left(\frac{\max(Y(i,j))}{RMSE}\right) \dots\dots\dots(5)$$

Here $i = 1,....,M$ and $j = 1,....,N$.
The RMSE (Root Mean Square Error) is calculated as:

$$\left(\frac{\sum_{i=1}^{M} \sum_{j=1}^{N} (X(i,j)-Y(i,j))^2}{M \times N}\right)^{\frac{1}{2}} \dots\dots\dots(6)$$

From eq. (5) and (6) it is clear that the higher values of PSNR as output will improve the contrast of the output images.

**II. (g) Absolute Modified Mean Brightness Error (AMMBE)**

The AMMBE can be calculated as $AMMBE(X,Y) = |X_{MM} - Y_{MM}|$ ………………………………………..(7)

The smaller value of AMMBE shows the brightness preservation of output image $Y$, with respect to input image $X$ [7].

### III. PROPOSED APPROACH

Our proposed methodology applies well to the low contrast medical images. Using hyper-kurtosis ($\beta$), it enhances image contrast and preserve the brightness. After calculating the mean of an image and considering $\tau$ as 3, hyper-kurtosis of the image is computed. Here $\tau$ is taken as 3[15]. For hyper-kurtosis $< \tau$, the modified mean (MM) can be calculated by taking the square-root of the sum of the mean and hyper-kurtosis. For hyper-kurtosis $\geq \tau$, different values in the formula of power-law can generate different MM, obtained by putting different values of $\gamma$ within the range of [0,1]. A suitable value of MM can be chosen where the output is found satisfactory. Then the modified mean (MM) can be obtained by taking the square root of the addition value or subtraction value. Subsequently based on this particular point the histogram of the input image can be separated into two separate parts and over those parts Equalization technique can be applied separately. Finally the union of both separate divisions are taken for producing a far better contrast enhanced output image.

### IV. RESULTS AND DISCUSSIONS

The performance of the proposed method of HKMDHE is tested with 52 test samples. The input test CT scan images of human brain (low contrast) in different cross sections are compared with CLAHE images and the modified images. The sample CT scan image is shown below as an example.

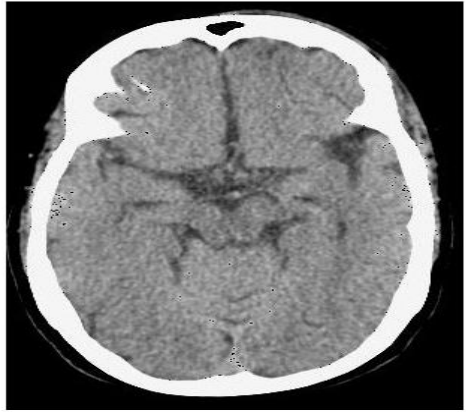

Fig1. CT scan image of human brain in different cross section

The plots of CLAHE and HKMDHE over CT scan images of human brain are shown in fig 2(a) and 2(b).

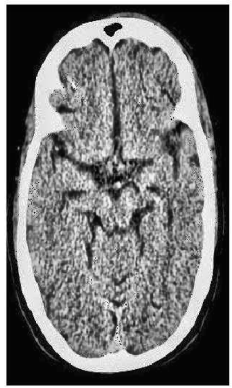
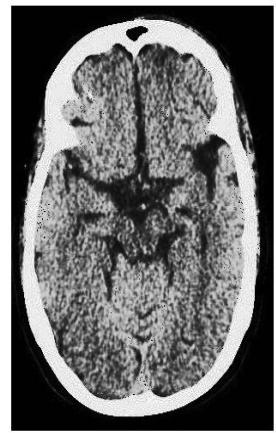

Fig 2(a)                                                                 Fig 2(b)

Fig-2 CT Scan Images of Human Brain after applying CLAHE (fig 2(a)) and HKMDHE (fig 2(b)) respectively

From the above plots, it is clear that after applying HKMDHE the contrast of CT scan images has increased substantially as compared to CLAHE.

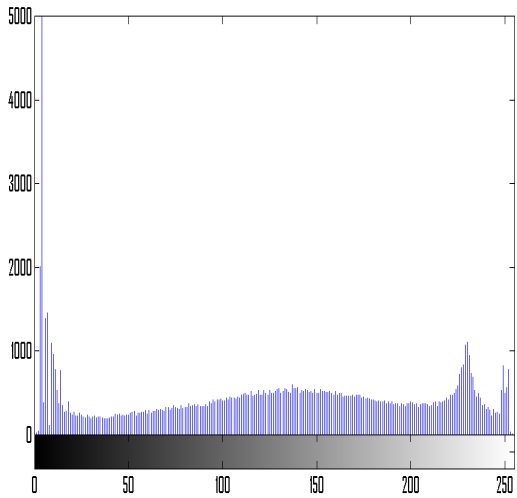 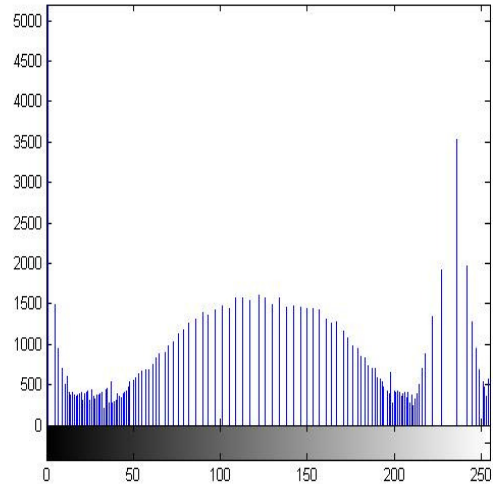

Fig 3(a)                                                        Fig 3(b)

Fig 3(a) & (b) - Histograms of Corresponding Images (X Axis: Pixel Intensity Values [0-255] & Y Axis: Number Of Pixels) by CLAHE (Fig 3(a)) and HKMDHE (Fig 3(b)) respectively.

From Fig.3(a) and Fig.3(b) it is clear that our proposed HKMDHE achieve more prominent standard normal distribution curve than CLAHE.

Comparison of PSNR and AMMBE values are shown in below table.

Table I:

|  | CLAHE | Proposed HKMDHE |
|---|---|---|
| PSNR | $17.92 \pm 0.2$ | $18.49 \pm 0.9$ |
| AMMBE | $0.10 \pm 0.009$ | $0.0125 \pm 0.01$ |

Hence, the contrast enhancement of input image is also better than CLAHE as the PSNR values are higher. The brightness of the input image is properly preserved in comparison with CLAHE as the AMMBE values are very small.

## V. CONCLUSION

The above results clearly show that our proposed method for contrast enhancement of medical images is much more efficient than the classical medical procedure like CLAHE. In future work we plan to apply signal processing tools like MFDFA, Wavelets, S-transform in contrast enhanced biomedical images for extracting more prominent features to characterize the normal and disease tissues.

## ACKNOWLEDGEMENT

S.Mandal is thankful to Bankura Sammilani Medical College and Hospital, Bankura, West Bengal, India for providing the CT images of human brain in different cross-section.